\documentclass[runningheads]{llncs}

\usepackage{eccv}

\usepackage{eccvabbrv}

\usepackage{graphicx}
\usepackage{booktabs}

\usepackage[accsupp]{axessibility}  

\usepackage{hyperref}

\usepackage{orcidlink}

\begin{document}

\title{HAVANA: Hierarchical stochastic neighbor embedding for Accelerated Video ANnotAtions} 

\titlerunning{HSNE for Accelerated Video Annotations}

\author{Alexandru Bobe \inst{1} \and
Jan C. van Gemert \inst{1}\orcidlink{0000-0002-3913-2786}
}

\authorrunning{A. Bobe and J.C. van Gemert}

\institute{Delft University of Technology, The Netherlands}

\maketitle

\begin{abstract}
  Video annotation is a critical and time-consuming task in computer vision research and applications. This paper presents a novel annotation pipeline that uses pre-extracted features and dimensionality reduction to accelerate the temporal video annotation process. Our approach uses Hierarchical Stochastic Neighbor Embedding (HSNE) to create a multi-scale representation of video features, allowing annotators to efficiently explore and label large video datasets. We demonstrate significant improvements in annotation effort compared to traditional linear methods, achieving more than a 10x reduction in clicks required for annotating over 12 hours of video. Our experiments on multiple datasets show the effectiveness and robustness of our pipeline across various scenarios. Moreover, we investigate the optimal configuration of HSNE parameters for different datasets. Our work provides a promising direction for scaling up video annotation efforts in the era of video understanding.
\end{abstract}

\section{Introduction}
\label{sec:intro}
The scarcity of labelled data continues to be an obstacle to progress in video understanding tasks for new domains. For instance, applications in underwater exploration \cite{moniruzzaman2017deep}, medical procedures \cite{khalid2020evaluation}, and autonomous driving \cite{wong2020mapping} are delayed due to the lack of high-quality data.

Even in established domains like surveillance \cite{subudhi2019big} and sports \cite{vinyes2017deep}, the quality of labelled data is not always on par with the requirements. For example, there is immense potential to enhance the analytics for tennis players, coaches and sports fans. Better strategies for players, personalized training programs for coaches, and increased audience engagement for fans would all be possible. However, the publicly available annotated tennis datasets are insufficient for these complex tasks \cite{hovad2024classification}.   

While recent years have seen remarkable advances in video understanding, the models for these tasks are still data-hungry \cite{kaushal2019learning,rangasamy2020deep}. Tasks like action recognition \cite{kong2022human}, temporal localization \cite{liu2022end}, and anticipation \cite{girdhar2021anticipative} rely on annotated datasets which are extremely labour-intensive to curate. 

\begin{figure}[t]
    \centering \includegraphics[width=0.65\linewidth]{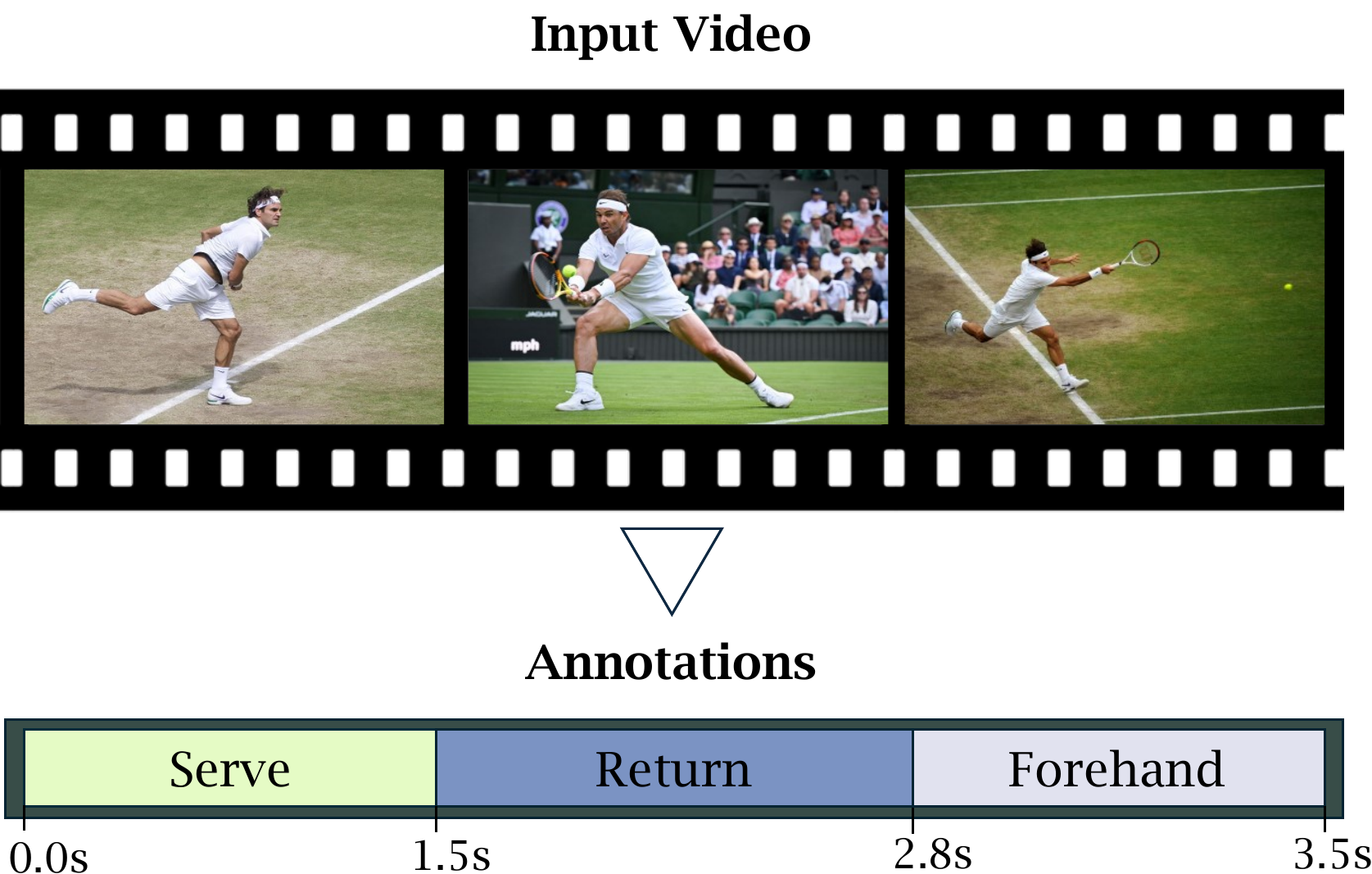}
    \caption{An example of temporal video annotation. The goal of this task is to create text annotations which identify the actions happening at each moment in the input video. These annotations can be further used to train new models or by domain experts for analysis.}
    \label{fig:ov}
\end{figure}

To curate these datasets, human annotators have to use annotation tools. The annotation tools take as input videos and human effort and output temporal annotations, as illustrated in Figure \ref{fig:ov}. Unfortunately, traditional annotation tools \cite{dutta2019via, shrestha2023feva} force human experts to label iteratively each video from start to finish, making the process unscalable and time-consuming.

A key challenge in visualizing large video datasets for annotation is what we call overflowing. Overflowing occurs when the number of data points to be visualized exceeds the limits of what can be meaningfully displayed in a 2D space. Traditional dimensionality reduction techniques struggle with this issue, leading to cluttered and uninformative visualizations that hinder efficient annotation.

In this paper, we challenge the status quo of the video annotation tools that are unscalable and time-inefficient. We create an effort-efficient and scalable annotation pipeline to accelerate temporal video annotations. Rather than forcing the human annotator to watch linearly each video, we exploit the similarities in videos to accelerate the annotation process in our pipeline.

Our pipeline takes as input pre-extracted features from any action recognition model. This is possible because our pipeline is model-agnostic and functions with any fixed-size feature type. Naturally, the quality of the features influences the annotation process. It is also possible to input in our pipeline frames instead of features. However, inputting frames is less efficient due to the increased storage requirements for frames compared to features, which results in longer processing time and worse visualisations.

After extracting and inputting the features in our pipeline, the user specifies some parameters for the Hierarchical Stochastic Neighbor Embedding (HSNE) \cite{pezzotti2016hierarchical}. We chose to use the HSNE technique for its ability to embed high-dimensional points into 2 dimensions. This means that features corresponding to similar actions are placed together, making it possible to annotate in bulk. Moreover, HSNE is scalable and continues working when presented with more data, solving the overflowing problem.

After HSNE has finished, a visual representation of the data is displayed. Using this initial visualization, the human annotator can explore deeper levels of detail in the hierarchical visualization using a selection tool. At the first level of the hierarchy, the annotator can use the selection tool again to choose groups of related points and efficiently input annotations for the entire selection.

Our contributions are as follows. We propose a novel, scalable annotation pipeline that uses the similarities between video frames to accelerate the annotation process. We compare our pipeline to the conventional approaches and quantify the improvement or explain why the other method cannot handle such large amounts of data. We use our pipeline in multiple scenarios, including popular datasets, to find the best approaches and observe the reliability and usability of the pipeline. 

\section{Related Work}
\label{sec:related}

\paragraph{Video Understanding.}
The expected outcomes of video understanding changed over time \cite{tang2023video, lavee2009understanding, huang2018makes}. Originally, the video understanding domain focused on foundational tasks like determining if an event has occurred and extracting an event summary \cite{lavee2009understanding}. Later, the field evolved into more intricate tasks, including captioning videos with descriptions \cite{abdar2023review}, answering questions about videos \cite{wang2023chatvideo} and anticipating the progression of the videos \cite{furnari2020rolling, tang2023video}. The field advancement in the complexity of tasks brought the need for more expressive models with increased levels of video interpretation \cite{borges2013video, wu2021towards} and sufficiently large datasets \cite{strafforello2023current}. However, these large datasets take a long time to be manually curated. Given the advancements in action recognition and temporal action localization, we believe the curation of the datasets can be sped up.


\paragraph{Temporal Action Localization.}
Temporal Action Localization (TAL) is a video understanding task that aims at splitting and categorizing the temporal intervals in untrimmed videos. Afterwards, it outputs each action's start and end time and the action category \cite{hu2024overview, ding2023temporal, warchocki2023benchmarking}. The most popular deep-learning techniques for TAL can be classified depending on the design method into anchor‑based methods \cite{shou2016temporal}, boundary‑based methods \cite{lin2019bmn}, and query-based methods \cite{liu2022end,hu2024overview}. Query-based methods are the most recent and naturally perform best when trained with large enough datasets \cite{zhang2022actionformer,warchocki2023benchmarking}. However, large annotated datasets are not available for some real-world specific actions, like tennis videos \cite{host2022overview} or network data \cite{guerra2022datasets}. 

\paragraph{Temporal Video Annotations.}
Temporal video annotation, also called event annotation \cite{shrestha2023feva}, is the process of marking temporal regions of interest in a video. The conventional method for tackling this task involves employing dataset-specific software entirely controlled by a human oracle \cite{damen2022rescaling, idrees2017thumos}. Moreover, the annotation of videos is linear as the human oracle has to annotate one video at a time. Imagine a person tasked with annotating numerous hours of video content. Each video must be watched entirely to create accurate annotations, requiring significant time and attention.

There exists general software for this task \cite{kipp2014anvil, dutta2019via, aubert2012advene, wittenburg2006elan, halter2019vian}. For example, VIA \cite{dutta2019via} is an open-source platform where users can annotate videos in multiple ways, including temporal annotations. The learning curve for the platforms is steep and the annotation process remains linear at best \cite{poorgholi2021t}. 

The progress in video understanding offers the opportunity for automating parts of the video annotation process. NOVA \cite{heimerl2019nova} brings semi-automation and explainability to the annotation process. However, the method does not solve the cold-start problem. The cold-start problem means that even the most efficient TAL models need huge amounts of data to perform satisfactorily. The performance is not presented in the paper \cite{heimerl2019nova} and we expect the gain in annotation speed to be neglectable for most tasks. FEVA \cite{shrestha2023feva} tries to solve the steep learning curve of annotation software for human oracles. Nevertheless, FEVA is still linear in terms of annotation time. t-EVA \cite{poorgholi2021t} introduces the possibility of 
better-than-linear annotation speed while keeping satisfactory accuracy. t-EVA uses a lasso tool on pre-extracted features embedded in a 2D space. While this approach offers several advantages, it still suffers from certain drawbacks. One of the drawbacks is the speed of creating the embedding \cite{pezzotti2016hierarchical}. Another drawback is the impossibility of visualizing a growing amount of features in the 2D space. Essentially, you're constrained by the size of your canvas, represented by the dimensions of your monitor. In our paper, the problems of time and dimension are solved by using pre-extracted features and a hierarchical dimensionality reduction technique.

\paragraph{Feature Extraction.}
Query-based methods for temporal action localization use features extracted using techniques from action recognition. For example, ActionFormer \cite{zhang2022actionformer} uses different visual features extracted with various backbones depending on the dataset. For the ActivityNet 1.3 dataset, \cite{ghanem2018activitynet} ActionFormer uses visual features from the R(2+1)D-34 model \cite{tran2018closer}. Moreover, for the EPIC Kitchens 100 dataset \cite{damen2022rescaling} ActionFormer uses visual features from SlowFast. It becomes evident that an effective annotation solution is agnostic to the underlying model and leverages various types of visual features.  

\paragraph{Dimensionality Reduction.}
Dimensionality reduction techniques can be classified into two categories: linear and non-linear methods \cite{reddy2020analysis}. Two of the most popular linear methods are PCA \cite{abdi2010principal} and LDA \cite{ye2004two}. Linear methods are widely used for their simplicity and efficiency. The main idea of these methods is to retain the most critical information from the original dataset.

As deep learning advances, the significance of image and video datasets has become paramount. Linear relations are not enough to deal with this complex data. Non-linear methods make it possible to reveal patterns in the data. t-Distributed Stochastic Neighbor Embedding (t-SNE) \cite{van2008visualizing} is a non-linear dimensionality reduction technique that preserves pairwise similarities between data points in the high and low-dimensional spaces. While t-SNE is versatile and applies to many use cases, it has significant limitations for our application.

A major drawback of t-SNE for video annotation is its difficulty in visualizing large datasets in a fixed 2D space. As the number of data points increases, the 2D visualization becomes cluttered and less informative, making it challenging for annotators to distinguish between different actions. This issue, which we call overflowing, occurs when the number of data points to be visualized exceeds the limits of what can be meaningfully displayed in a 2D space.

Moreover, the performance of t-SNE degrades quickly in terms of speed and visualization quality with larger datasets \cite{pezzotti2016hierarchical}. Given our motivation to improve annotation speed and handle large video datasets, this performance degradation is a significant drawback for our use case.

UMAP, another non-linear dimensionality reduction technique, promises to solve these issues. However, it was demonstrated that UMAP suffers from the same problems as the best-performing variants of t-SNE \cite{kobak2019umap}.

A technique called Hierarchical Stochastic Neighbor Embedding (HSNE) \cite{pezzotti2016hierarchical} addresses both the time performance issues and the overflowing problem. On the MNIST dataset \cite{lecun1998gradient}, HSNE performs more than ten times faster than t-SNE alone \cite{pezzotti2016hierarchical}. Additionally, HSNE, being a hierarchical technique, effectively tackles the 2D space limitations for displaying embeddings, thus solving the overflowing problem. These properties make HSNE suitable for our goal of creating a fast and scalable video annotation pipeline.

\section{The Annotation Pipeline}
\label{sec:method}

\begin{figure}[t]
 \center
  \includegraphics[width=0.8\columnwidth]{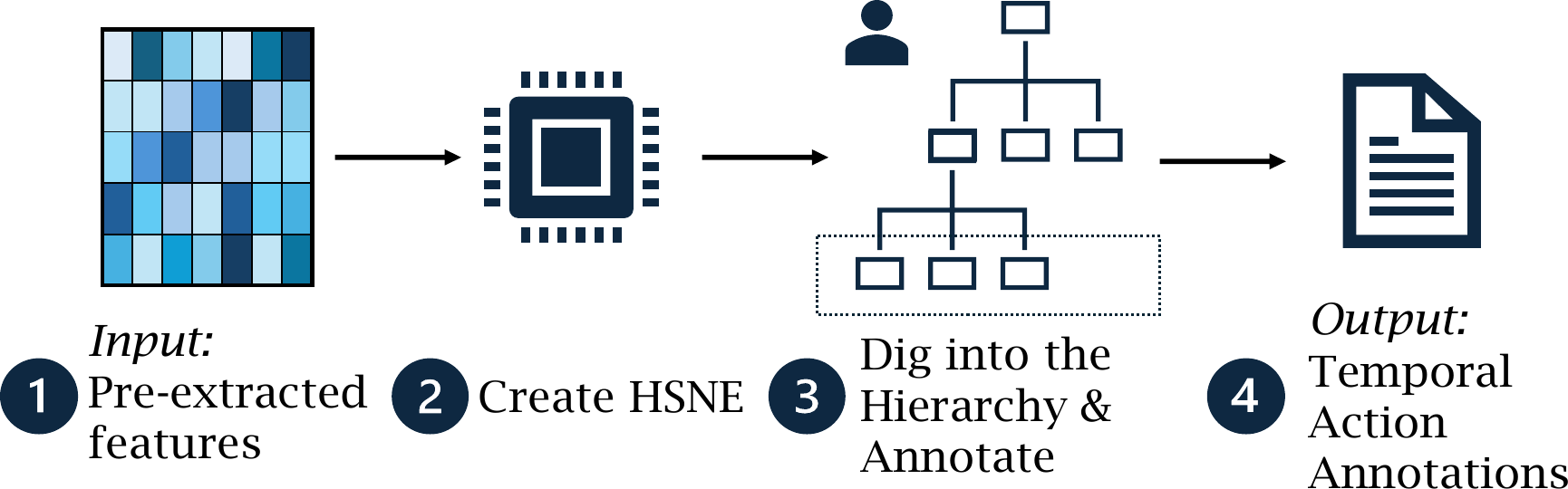}
  \caption{The figure depicts an overview of our annotation pipeline. (1) The model-agnostic pre-extracted features are used as input in the analysis. (2) The HSNE (Hierarchical Stochastic Neighbor Embedding) analysis is created. (3) Human-in-the-loop approach for digging into the hierarchy and annotating at the deepest scale. (4) Temporal Action Annotations are outputted in JSON format. }
  \label{pipeline}
\end{figure}

Here, we present and motivate the components of our annotation pipeline. Figure \ref{pipeline} gives an overview of the pipeline. The pipeline, which follows a human-in-the-loop approach, takes as input extracted features and outputs temporal action annotations, ready to be visualised and further refined in open-source software like VIA \cite{dutta2019via}. In addition, our approach can handle features from trimmed and untrimmed videos, making it suitable for real-world applications.  

\subsection{Feature Extraction}
To get the best performance of our tool, pre-processing the videos for feature extraction has to be done. Video frames are also accepted as input, but the speed and accuracy decrease considerably. The main reasons for preferring features over frames are the size and the accuracy.  Extracted features convey several frames' information in a single feature vector, usually of size 2048, making it easier to process. On the other hand, a single RGB frame of size 320x180 takes approximately $172\,800$ values. The increased data size when using plain frames impacts the performance of the dimensionality reduction algorithm and constrains the number of videos that can be processed simultaneously. 

Our pipeline is feature-agnostic. This means that any video features can be used, as long as they have a fixed length. Usually, the features come from pre-trained action recognition models, like two-stream I3D \cite{carreira2017quo}, R(2+1)D \cite{tran2018closer} and SlowFast \cite{feichtenhofer2019slowfast}. 
Naturally, the features' quality and the dataset influence the dimensionality reduction algorithm, and implicitly the pipeline performance. 

\subsection{t-distributed Stochastic Neighbor Embedding (t-SNE)}

t-distributed Stochastic Neighbor Embedding (t-SNE) is a non-linear dimensionality reduction technique for visualising high-dimensional data in a lower-dimensional space \cite{van2008visualizing}. In our annotation pipeline, we use t-SNE to visualise the high-dimensional features on the 2-dimensional screen. More specifically, any time the user wants to visualise a subset of the points, t-SNE creates a 2d embedding. Here, we give an overview of how t-SNE creates this embedding. For a more detailed explanation of the original method, we refer to the original work \cite{van2008visualizing}, whereas for how t-SNE works in detail inside HSNE, we refer to this paper \cite{pezzotti2016hierarchical}.     
The main steps of t-SNE are:
\begin{enumerate}
\item Compute pairwise similarities between all high-dimensional points using a Gaussian kernel.
\item Transform pairwise similarities into joint probabilities by normalizing the similarities for each data point.
\item Define a similar set of joint probabilities in the low-dimensional space and optimize the positions of low-dimensional points.
\item Visualize the data by plotting the low-dimensional embedding.
\end{enumerate}

t-SNE is a powerful and versatile technique, however, it cannot handle the overflowing problem alone. More specifically, t-SNE cannot embed large datasets with too many points. Therefore, we employ Hierarchical Stochastic Neighbor Embedding to solve this problem. 

\subsection{Hierarchical Stochastic Neighbor Embedding (HSNE)}

Hierarchical Stochastic Neighbor Embedding (HSNE) is a dimensionality reduction technique. It is an SNE technique and solves the problem of speed and space required to visualize large datasets. Here, we give an overview of how the method works. For a more in-depth explanation, we refer to the original work which introduces the technique and provides an implementation \cite{pezzotti2016hierarchical}. 

\begin{figure}[t]
 \center
  \includegraphics[width=0.4\columnwidth]{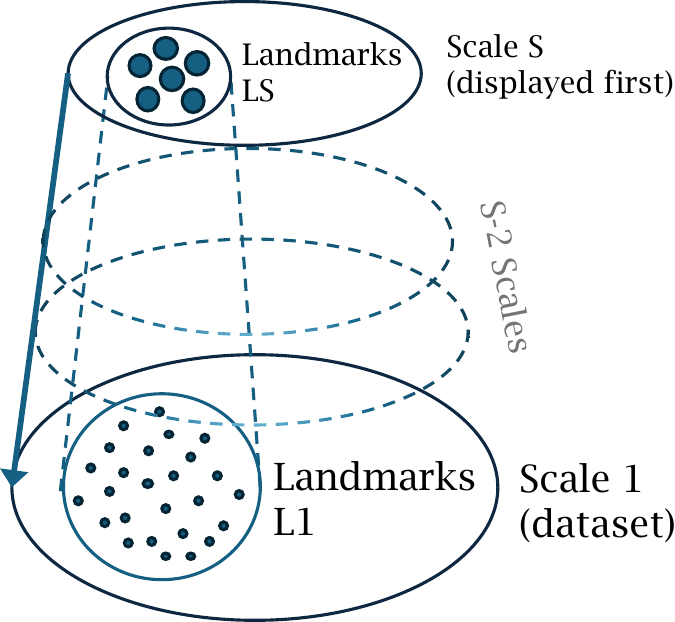}
  \caption{A visualisation of how the landmarks and scales work in HSNE. The area of influence of the landmarks in Scale S can be seen in Scale 1. The number of intermediate scales varies depending on the user-specified S parameter.}
  \label{landmarks}
\end{figure}

The core concept of HSNE involves operating across multiple scales or levels, denoted by the user-specified parameter \textit{S}, rather than embedding all high-dimensional data points into a single 2-dimensional scale. The algorithm identifies landmarks at each scale and utilizes t-SNE to project them into a 2D space for visualization. Figure \ref{landmarks} describes the intuition behind how scales and landmarks work in HSNE.

Intuitively, the main steps of the HSNE method are: 
\begin{enumerate}
\item The Euclidean distances between the high-dimensional data points are computed. The distances are used to calculate each point's k-nearest neighbourhood (KNN) and create a KNN graph.
\item The KNN graph is used to select the landmarks or points in the next scale.
\item For each landmark, an area of influence over the points in the previous scale is computed. 
\item Overlaps in the areas of influence are used to create similarities between the points at the new scale. Steps 3 and 4 are repeated to create landmarks for each scale.
\item Similarities are used on request to create an embedding using t-SNE. The embedding is used for annotating when HSNE is integrated into the annotation platform.
\end{enumerate}

\subsection{Annotation Platform}

The annotation platform is built on top of the Python wrapper of the HSNE implementation \cite{pezzotti2016hierarchical}. As in the original implementation, the user can select the number of scales and iterations for each t-SNE analysis and optionally input text labels to improve the visualisation. 

The implementation was modified to visualize frames while hovering over points in the HSNE analysis. For each feature vector, a representative frame is pre-extracted from the video. This does not influence the HSNE algorithm but facilitates the annotation process. Moreover, keyboard shortcuts were added to enable the annotation process, using a lasso tool and a pop-up window for text input. 

\section{Experiments}
\label{sec:experiments}

Our video annotation pipeline leverages pre-extracted features and dimensionality reduction to accelerate and enhance the process of creating temporal action annotations. Through experiments, we aim to empirically demonstrate the pipeline's effectiveness on real-world video data, investigate the impact of the advanced dimensionality reduction technique and compare the pipeline's performance against existing linear annotation tools. 

We conducted experiments using features extracted with various techniques from different datasets to evaluate the pipeline's performance across multiple scenarios.

\subsection{Datasets \& Features Used}
We used features extracted from three datasets throughout the experiments: a synthetically generated dataset, Thumos14 \cite{idrees2017thumos} and Epic-Kitchens-100 \cite{damen2022rescaling}.

\paragraph{Synthetic Features.}\label{sf} 
The synthetic dataset was created to investigate the pipeline's performance under ideal conditions with perfect features.
We created the synthetic data to match the class distribution of the THUMOS14 test set. This process led to the ratio between \textit{Background} and \textit{Actions} to be 2:1.
To design the features, we created a one-hot encoding for the 21 classes present in THUMOS14. The one-hot encodings mimicked the ground truth label of the features extracted from the THUMOS14 test set. Afterwards, we added Gaussian noise to the features. Gaussian noise was created by sampling 21 times for each feature vector from a Gaussian distribution with the mean at 0 and a standard deviation of 1.

\paragraph{THUMOS14 Features.}
THUMOS14 is a large-scale dataset for temporal action localization in untrimmed videos, with multilabel videos from 20 sport action classes.
We used the features included in the ActionFormer paper \cite{zhang2022actionformer}, extracted from the THUMOS14 test set. The features were extracted from two-stream I3D models \cite{carreira2017quo} pre-trained on Kinetics \cite{carreira2017quo}, utilizing 16-frame clips at 30 fps and a stride of 4 frames. This configuration gave a single feature vector every 0.1333 seconds.
The total amount of videos used for extraction was 213, which amounted to roughly 12 hours of videos. The ratio between \textit{Background} and \textit{Actions} in the videos corresponding to the features was 2:1.


\paragraph{Epic-Kitchens-100 Features.}
EPIC-Kitchens-100 is a challenging egocentric video dataset captured from wearable cameras in kitchen environments, containing 97 verb classes. 
We used the features from the ActionFormer paper \cite{zhang2022actionformer}, extracted from the EPIC-Kitchens-100 validation set. We chose to use the validation set, as the test set's labels were unavailable. The features were extracted using the SlowFast model pre-trained on the training set of EPIC Kitchens 100 for action classification. This process involved utilizing 32-frame clips at a frame rate of 30 fps with a stride of 16 frames. Therefore, the process yielded a singular feature vector approximately for every 0.5333 seconds of videos. The total amount of videos used for extraction was 138, which amounted to more than 13 hours of videos. The ratio between \textit{Background} and \textit{Actions} in the videos corresponding to the features was 1:2. The ratio shows how action-dense the dataset is compared to THUMOS14.

%



\subsection{Exp 1: Annotation Effort Improvement} 

Temporal video annotations are time-consuming and require a lot of human effort. In this experiment, we investigate how much effort users can save using our pipeline compared to traditional annotation methods. How much improvement in annotation effort can be achieved compared to conventional annotation methods? To answer this question, we estimated the effort needed to annotate videos using our pipeline and a conventional method, the VIA tool \cite{dutta2019via}.

To estimate the effort needed for annotating with VIA we used the ground truth labels of the videos from the THUMOS14 test set. We assumed that for every action segment we wanted to annotate a button had to be clicked once. This is a lower bound, as we have seen in our experience that the VIA tool requires multiple clicks to adjust the annotated segment in the desired way. Moreover, VIA is a linear method. The linearity of VIA means that each video has to be annotated separately and no speed-up can be achieved.

To estimate the effort needed for annotating with our tool we used the ground truth labels of the features from the THUMOS14 test set. The interaction with the system had to be automated. To automate the drilling part of our pipeline, we used the Agglomerative Clustering algorithm with the \textit{Single} linkage criterion \cite{sibson1973slink} from the Scikit-Learn library \cite{scikit-learn}. 

The \textit{Single} linkage criterion defines the distance between two clusters as the minimum of the distances between all pairs of elements. We chose this technique for its ability to create uneven cluster sizes, suitable for our imbalanced classes in the dataset. In this experiment, the \textit{Background} pseudo-class played a role in unbalancing the clusters' sizes. 

Moreover, \textit{Agglomerative Clustering - Single linkage} works well on non-globular data, which is the case for us. We used the expected number of distinct labels present to choose the number of clusters. To mimic the drilling process of a human in the HSNE analysis, we had to choose a tree traversal algorithm. We went with the Depth-First Search (DFS), assuming that is how a human would use the system.

\begin{figure}[t]
 \center
  \includegraphics[width=0.6\columnwidth]{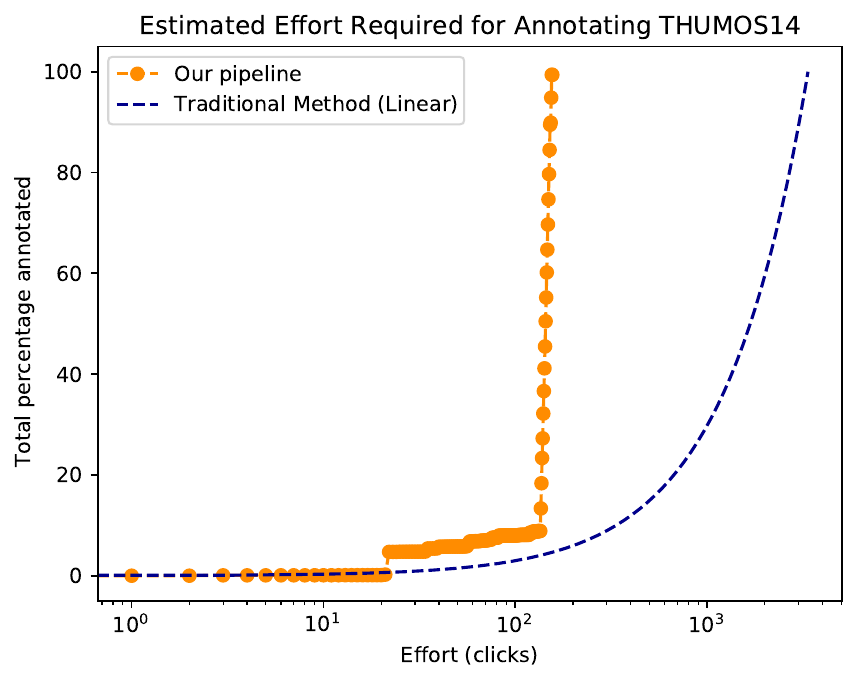}
  \caption{Estimated effort in clicks needed to annotate the test set of THUMOS14 using our pipeline and a traditional linear method. The results show a more than 10 times improvement when using our method.}
  \label{Effort}
\end{figure}

Results in Figure \ref{Effort} show a significant improvement in estimated effort when using our pipeline compared to a traditional linear method. The figure shows an estimate of how many clicks are needed to annotate the test set of THUMOS14 with both methods. The \textit{Total Percentage Annotated} represents the mean percentage of each class annotated.  

The jump in our method at around 100 clicks is attributed to the uneven cluster sizes created by the Agglomerative Clustering and uneven class distribution in the test set. This claim was verified by manually annotating a subset of the data. We conclude that our method shows more than a 10x improvement in effort when annotating more than 12 hours of videos.   

\subsection{Exp 2: Pipeline Performance \& Feature Quality}

As the pipeline takes features as input, the quality of these features inherently influences the pipeline's performance. In this experiment, we observe how the pipeline would work with perfect synthetic features. This way, we can answer the question: What is the best achievable performance using our pipeline, and how does the quality of input features influence the annotation process?


When using the perfect features the different classes and the background are perfectly separated from the last scale. In the case of THUMOS14 features, the different classes are mostly separated, however, the \textit{Actions} and \textit{Background} are not yet separated at this scale. Since these classes are not separated from the last scale, the human annotator must do more work throughout the scales to annotate.

\begin{figure}[t]
 \center
  \includegraphics[width=0.6\columnwidth]{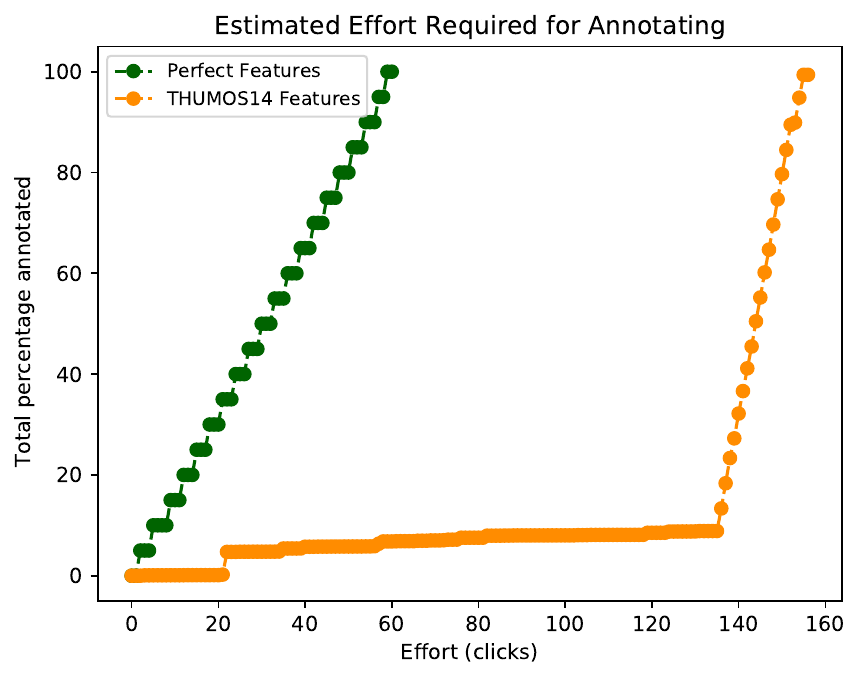}
  \caption{ An estimation of effort required to annotate the test set of THUMOS14. Perfect features represent the synthetic features, created as explained in \ref{sf}. The THUMOS14 features correspond to the test set features of the THUMOS14 dataset. The plot shows a 50\% possible improvement in terms of effort achieved by increasing the feature quality. }
  \label{perf_feats}
\end{figure}

Figure \ref{perf_feats} shows an upper bound in the pipeline's performance. This performance could be achieved just by improving the features' quality in terms of distinguishing between different action classes and between background and action in the videos. We estimate a 50\% reduction of effort when using perfect features compared to the features we used for the THUMOS14 test set. In summary, higher-quality input features that can effectively separate different action classes and backgrounds from actions have the potential to significantly improve the performance of the annotation pipeline and reduce the effort of the human annotator.

\subsection{Exp 3: Impact of Landmark Selection}

A significant part of the HSNE analysis is selecting meaningful landmarks at each scale. In this experiment, we explain how the landmark selection process works in our pipeline. Moreover, we answer the question: does the landmarks selection of Hierarchical Stochastic Neighbour Embedding (HSNE) impact the pipeline, and how does it compare to simpler approaches?

In HSNE, the landmarks are the subset of data points selected and displayed at each scale of the hierarchical embedding. These points have to be representative of the global structure and density patterns of the high-dimensional data. HSNE identifies landmark points as those that have a high number of neighbours with other points. This allows HSNE to select landmarks that are central to dense data clusters and to avoid choosing outliers as landmarks.

\paragraph{Uniform sampling.} To assess how the selection method used by HSNE affects our pipeline, we replace this selection procedure with a baseline, uniform sampling. Then, we compare the effort estimations when annotating the THUMOS14 test set. The uniform sampling strategy follows the hierarchical structure of HSNE. In a 3-scale analysis with uniform sampling, the ratio of landmarks is 1:25:125. This means that each landmark at \textit{Scale 3} maps to 25 landmarks at \textit{Scale 2} and to 125 landmarks at \textit{Scale 1}.  

\begin{figure}[t]
 \center
  \includegraphics[width=0.6\columnwidth]{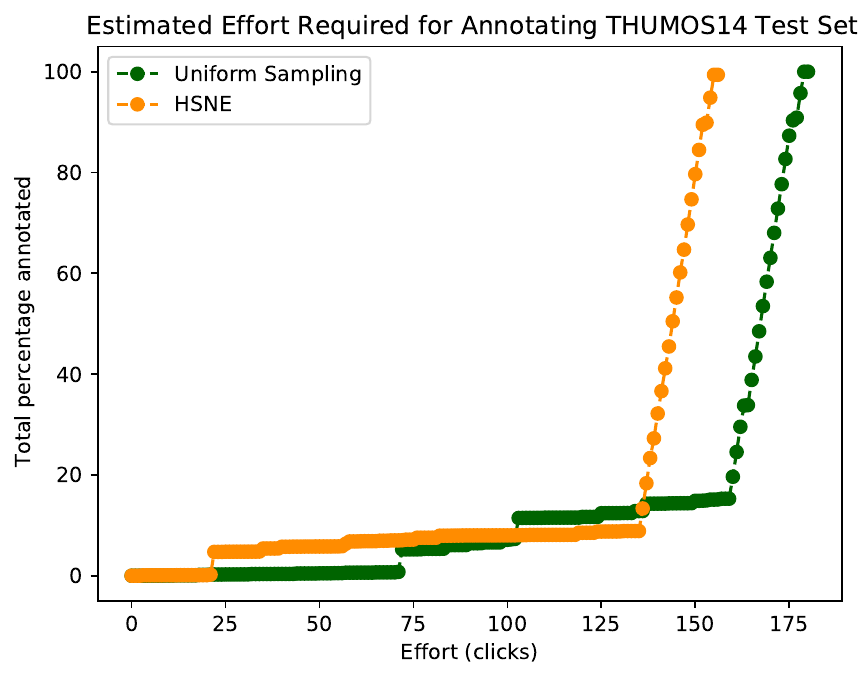}
  \caption{ The estimated effort required to annotate the THUMOS14 test set using extracted features. \textit{Uniform sampling} represents the baseline landmark selection strategy, where each point at \textit{Scale 3} maps to 125 points at \textit{Scale 1}. The plot shows a 13\% improvement when using the HSNE landmark selection strategy.}
  \label{uni_eff}
\end{figure}

Figure \ref{uni_eff} presents an effort estimation of annotating the THUMOS14 test set. In this estimation, the only variable that changed was the landmark selection strategy. The plot shows that the landmark selection strategy used by HSNE brought a 13\% improvement in effort compared to the uniform sampling strategy on the THUMOS14 test set. Moreover, HSNE's landmark selection strategy helps the human annotator understand the data better by displaying informative landmarks at each scale. Therefore, the effort required from the human annotator would be smaller thanks to HSNE. 

\subsection{Exp 4: Impact of Displayed Points}

After the drilling in the hierarchy has been done, the final step in the annotation pipeline is to annotate the landmarks in the first scale. To do this, the human annotator has to draw using a lasso tool. Intuitively, the difficulty of this process depends on multiple factors, including the dataset, the number of landmarks displayed and the level of granularity expected for the annotations. Out of these factors, the only factor we can influence is the number of displayed points by our pipeline. Naturally, we want to answer the following questions: How does the number of points displayed on the screen affect the annotation experience, and what is the recommended setting for two specific datasets?

The number of displayed landmarks on the first scale is based on the total number of points and how the drilling was performed. Separating clusters while drilling results in fewer points displayed at once at the first scale. However, the drilling process also takes cognitive effort from the human annotator. Therefore, if we can find a desirable range for the amount of landmarks displayed, we can automate the drilling process and save effort.  

We start this experiment by manually drilling and annotating 25 times. We then try to empirically find the right amount of points to be displayed for the THUMOS14 and Epic-Kitchens-100 datasets using automatic drilling. 

The automatic automatic drilling was performed using KMeans clustering, from the Scikit-Learn library \cite{scikit-learn}. We chose this method for its ability to create regular clusters, mimicking a basic approach taken by a human annotator. KMeans takes the number of clusters to find as a parameter. We vary this number based on how many displayed points we want to have at the first scale on average. 

\begin{figure}[t]
 \center
  \includegraphics[width=0.7\columnwidth]{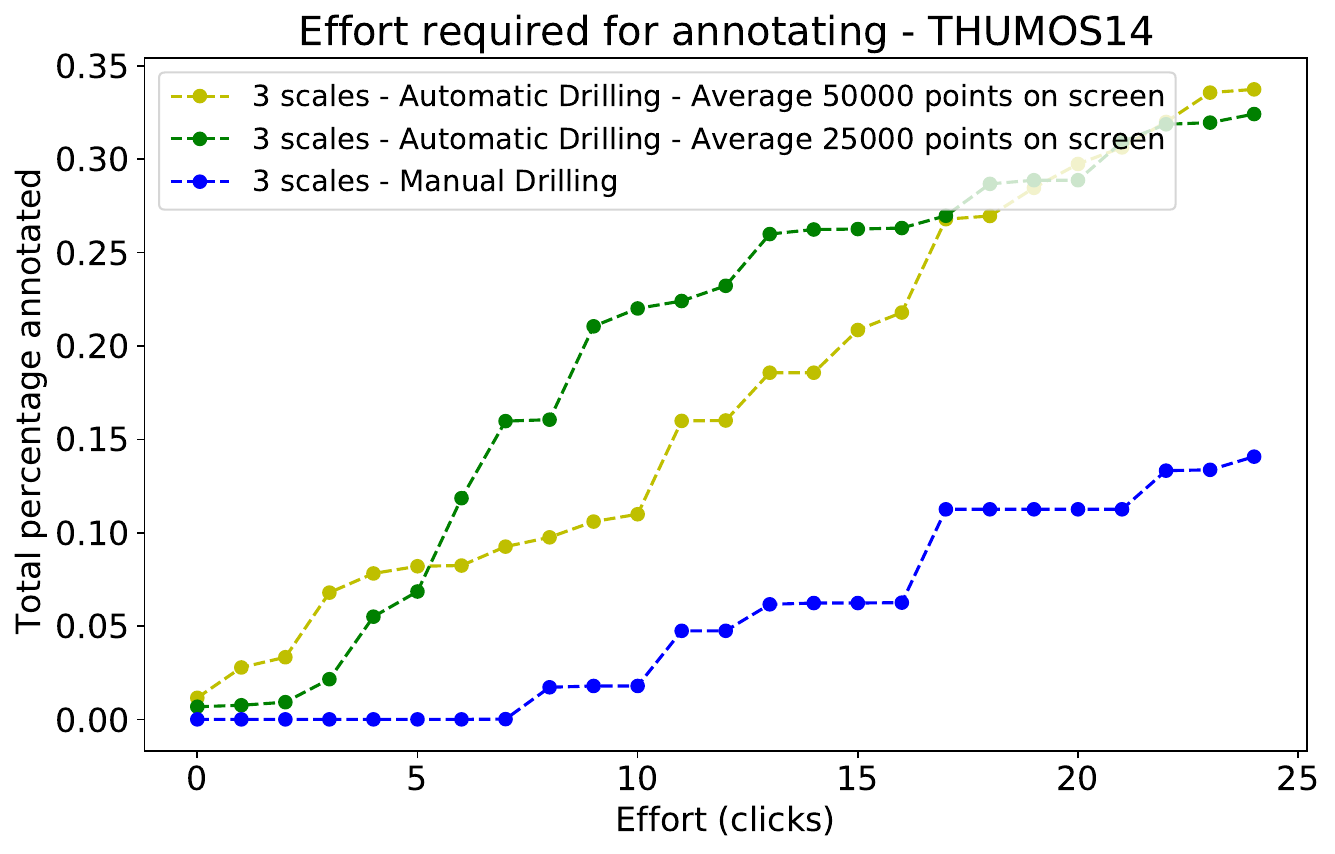}
  \caption{The total percentage we could annotate with 25 clicks when using manual and automatic drilling on THUMOS14. It is visible that automatic drilling can save effort.}
  \label{display_th}
\end{figure}

Figure \ref{display_th} shows how much percentage we were able to annotate in 25 steps using THUMOS14. We found that on average 25000 to 50000 landmarks was the right amount of displayed points for this dataset. We also tried more than 50000 landmarks displayed on average and the annotation process became infeasible. Moreover, we found that drilling can be done automatically, without affecting the annotation process. This way we could save effort and annotate more with the same amount of clicks.   

\begin{figure}[t]
 \center
  \includegraphics[width=0.7\columnwidth]{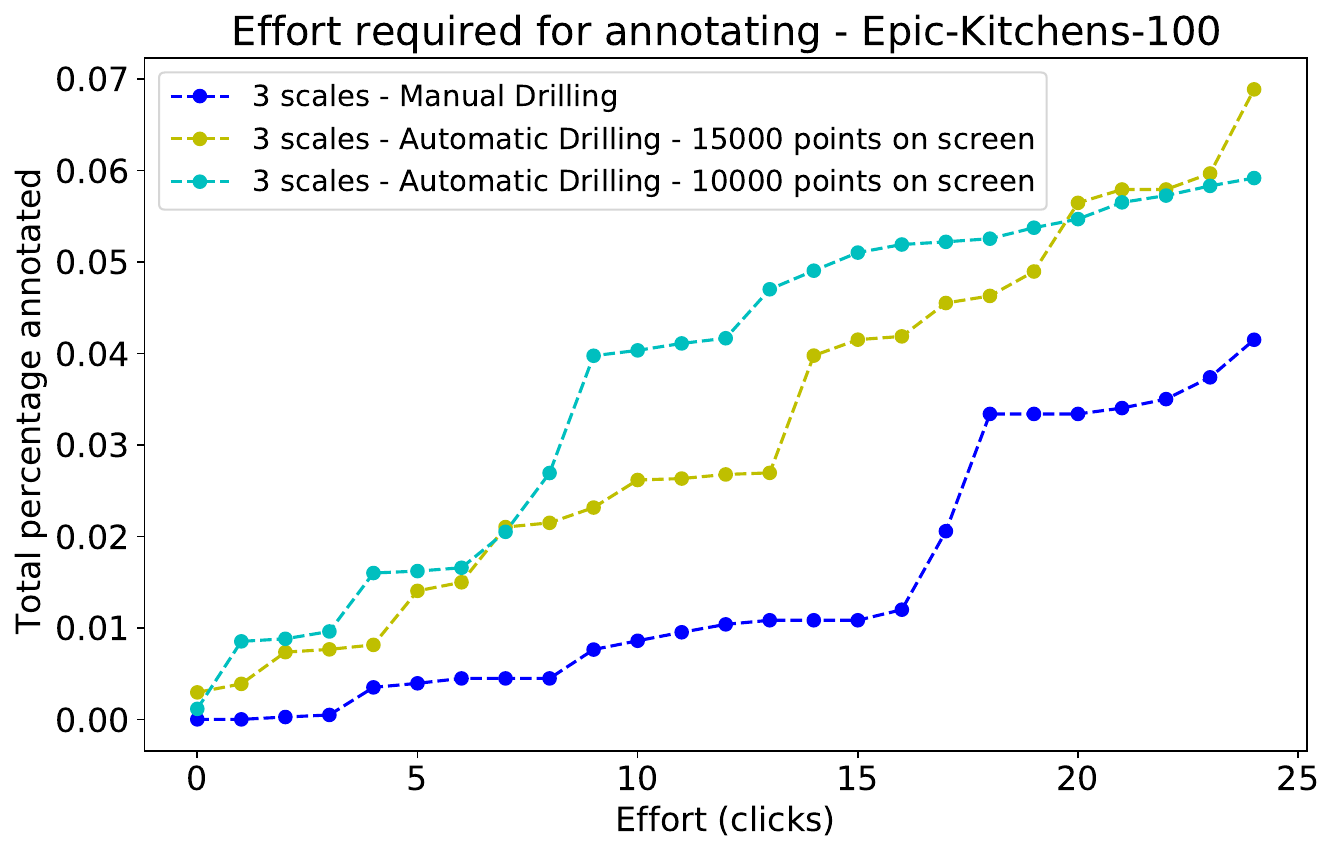}
  \caption{The total percentage we could annotate with 25 clicks when using manual and automatic drilling on Epic-Kitchens-100. It is visible that automatic drilling can save effort.}
  \label{display_ek}
\end{figure}

To verify that the automatic drilling works, we annotated again for 25 clicks on Epic-Kitchens-100. We found the range of displayed points for this case to be between 10000 and 15000. The results are presented in figure \ref{display_ek}. We cannot compare the results between Epic-Kitchens-100 and THUMOS14 as the datasets' difficulty and the features' quality are completely different. Nevertheless, we can conclude that automatic drilling works on both datasets and can save effort without impacting the annotation process.

\section{Conclusion}
\label{sec:conclusion}

Our proposed video annotation pipeline demonstrates significant potential for accelerating temporal action annotations. By using pre-extracted features and hierarchical dimensionality reduction, we achieve more than 10 times improvement in annotation effort compared to traditional annotation methods. Our experiments across multiple datasets highlight the effectiveness of our approach when used on datasets with multiple similar videos and provide insights into optimizing the pipeline for different scenarios.

One limitation of our annotation pipeline is the dependency on the quality of pre-extracted features. Additionally, the need for manual tuning of pipeline parameters for optimal results in different scenarios can be time-consuming and may require expert knowledge. Furthermore, our current evaluation is based on the number of clicks, as an approximation of the cognitive effort necessary for the annotator to complete the task.

Future work directions should include conducting a user study to evaluate the usability of the annotation pipeline and quantify the improvement in terms of annotation quality and cognitive effort compared to traditional annotation methods. Afterwards, the annotation pipeline can be used to create new datasets and eventually automatically adapt the pipeline's parameters based on scenario characteristics.

To conclude, our work provides a promising foundation for addressing the challenge of efficiently annotating large-scale video datasets. As video understanding tasks continue to evolve and demand larger, more diverse datasets, scalable annotation methods like ours will play a crucial role in advancing the field and its real-world applications.


%
%
\bibliographystyle{splncs04}
\bibliography{main}
\end{document}